\pdfoutput=1

\documentclass[11pt]{article}

\usepackage{acl}

\usepackage{times}
\usepackage{stfloats}

\usepackage{latexsym}
\usepackage{amsmath}
\usepackage{graphicx}
\usepackage{makecell}
\usepackage{listings}
\usepackage{xcolor}
\usepackage{hyperref}
\usepackage{fontawesome5}
\usepackage[T1]{fontenc}
\usepackage[utf8]{inputenc}
\DeclareUnicodeCharacter{00AD}{}

\usepackage{microtype}

\title{Directional Attractors in LLM Reasoning: How Similarity Retrieval Steers Iterative Summarization Based Reasoning}

\author{Charbel Barakat \\
  McGill University \\
  charbel.barakat@mail.mcgill.ca \\\And
  Luis Joseph Luna Limgenco \\
  McGill University \\
  luis.limgenco@mail.mcgill.ca \\\AND
  Cagatay Tekin \\
  McGill University \\
  cagatay.tekin@mail.mcgill.ca\\
}

\begin{document}

\maketitle

\begin{abstract}
Iterative summarization based reasoning frameworks such as InftyThink enable long-horizon reasoning in large language models (LLMs) by controlling context growth, but they repeatedly regenerate similar reasoning strategies across tasks. We introduce \emph{InftyThink with Cross-Chain Memory}, an extension that augments iterative reasoning with an embedding-based semantic cache of previously successful reasoning patterns. At each reasoning step, the model retrieves and conditions on the most semantically similar stored lemmas, guiding inference without expanding the context window indiscriminately. Experiments on MATH500, AIME2024, and GPQA-Diamond demonstrate that semantic lemma retrieval improves accuracy in structured domains while exposing failure modes in tests that include heterogeneous domains. Geometric analyses of reasoning trajectories reveal that cache retrieval induces directional biases in embedding space, leading to consistent fix (improve baseline accuracy) and break (degradation in baseline accuracy) attractors. Our results highlight both the benefits and limits of similarity-based memory for self-improving LLM reasoning.
\end{abstract}

\vspace{0.5em}
\begin{center}
\small
\faGithub\ \textbf{Code:} \url{https://github.com/cagopat/InftyThink-with-Cross-Chain-Memory}
\end{center}
\vspace{1em}

\section{Introduction}

Large Language Models (LLMs) currently face significant challenges when tackling complex, multi-step problems primarily due to the quadratic computational scaling of attention mechanisms. While frameworks like InftyThink have introduced iterative summarization-based reasoning to recursively compress and retain context, the compressed context they provide for each subsequent question does not retain the entirety of the details involved in solving previous questions.

This paper addresses the need for a more efficient and dynamically self-improving reasoning mechanism by integrating an embedding-based semantic cache into the InftyThink framework. We propose a system—InftyThink with Cross-Chain Memory—that algorithmically stores frequently used reasoning steps ("lemmas" or "tricks") from previously solved problems and retrieves the $k$ most semantically similar ones to augment the context window of a new problem. This approach aims to mitigate context-window pollution and reduce the repeated derivation of accurate reasoning steps by providing only maximally relevant lemmas to guide the model's reasoning.

Our principal contribution is within our \textbf{System Design (InftyThink + Cross-Chain Memory):} We propose and implement an extension of the InftyThink framework that incorporates an embedding-based semantic cache using the BGE model and cosine similarity. This mechanism provides a self-improving collective memory that dynamically augments the LLM's context window with reusable reasoning patterns from past inference tasks, moving beyond static prompt engineering. Through experiments on benchmark datasets, we quantify the trade-off between cache quality, size, and system performance, providing essential guidance for developing self-improving LLM reasoners.

\section{Related Works}
Prior work has explored several complementary approaches to improving reasoning in large language models under context and compute constraints. InftyThink \cite{inftythink} introduces an iterative summarization-based reasoning framework that enables long-horizon reasoning by recursively compressing intermediate steps, avoiding unbounded context growth. Related iterative refinement approaches such as Editor Chain \cite{editorchain} and Self-Refine demonstrate that repeated reasoning and editing cycles are more effective than single-shot prompting for complex tasks. Separately, work on context window pollution shows that injecting irrelevant or weakly related information into the context can significantly degrade model performance, even when the total context length increases \cite{contextpollution}. Research on analogical reasoning argues that problem solving in LLMs is fundamentally driven by the reuse of abstract strategies or “tricks” derived from similar problems \cite{yasunaga2023}, while recent geometric analyses of representation space suggest that reasoning outcomes and failure modes may correspond to structured directions in embedding space \cite{marks2023geometry}. Finally, studies on self-improving reasoners emphasize that reasoning behavior—such as strategy reuse and reflection—plays a critical role in downstream performance \cite{cognitivebehaviors}. In contrast to these approaches, which either rely on internal regeneration of strategies or unconstrained context expansion, our work integrates an embedding-based semantic cache into an iterative reasoning framework, selectively retrieving previously useful reasoning patterns to guide inference while explicitly examining the trade-offs between relevance, cache size, and domain heterogeneity.


\section{Data and Environment}
\subsection{Data}
The datasets used for this study are the same datasets used in \cite{inftythink}. Our first dataset is \textbf{MATH500}, which consists of 500 math problems sourced from renowned high school math competitions. Next is \textbf{AIME2024}, which is a dataset containing 30 math problems from the 2024 American Invitational Mathematics Examinations \cite{aime2024}. The problems in this dataset are considered highly challenging, as the examination is invitation-only \cite{aime2024}. The problems used in this dataset are not meant to be solved using basic computational methods and all have a specific integer solution between 000 and 999. Finally, we use \textbf{GPQA Diamond}, which is a 198 question dataset with graduate-level multiple-choice questions sourced from domain experts in the sciences \cite{gpqa2023}. The problems used in this dataset are categorized into three scientific categories including Biology, Physics, and Chemistry and provide a broader scope of questions to test our embedding-based approach for augmenting the context window.

\subsection{Environment}
To conduct experimental work, the OpenAI API Python client was used to format and issue API calls to query LLMs. Since our evaluated model was from the Qwen LLM family, we generated an API key through Alibaba Cloud in order to access the required models. The \textit{Transformers} and \textit{Datasets} libraries by HuggingFace were used to fetch the benchmarks, and a local BGE embedder was used to perform cache retrieval. The use of an API-based environment enabled experiments without requiring higher-order computing resources, permitting local testing through Jupyter Notebooks. 

\section{Methods}

The hypothesis is that lemma retrieval based on question-lemma similarity improves analytical reasoning performance by augmenting the context window with appropriate guidelines for approaching the question. The baseline model is vanilla InftyThink, which implements iterative summarization-based reasoning with a fixed token limit per reasoning iteration. It was implemented via prompts and separate API calls rather than a fine-tuned model. This approach was followed since the main goal of the paper is to achieve an algorithmic improvement rather than to establish a new state-of-the-art benchmark.

\begin{figure}[h]
    \centering
    \includegraphics[width=1\linewidth]{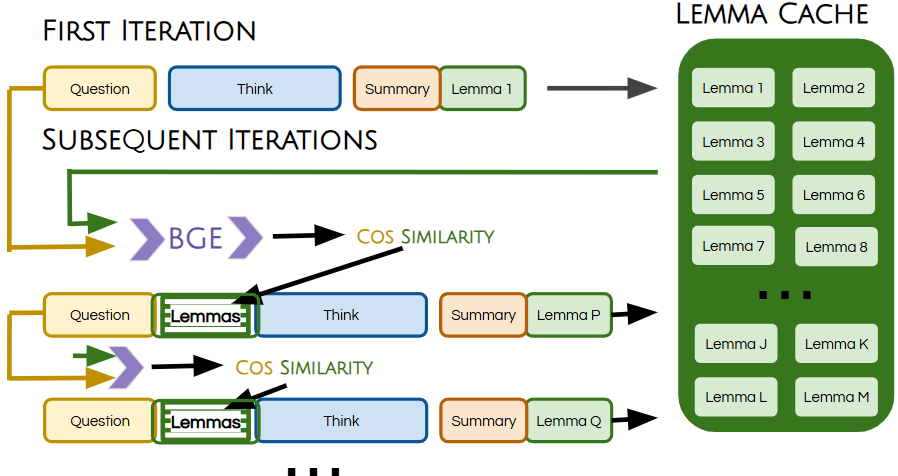}
    \caption{Lemma Augmented InftyThink}
    \label{fig:placeholder}
\end{figure}

\begin{table*}[h]
\centering
\begin{tabular}{lcccccc}
\hline
\textbf{Dataset} & \textbf{n} & \textbf{Vanilla (Baseline)} & \textbf{Cache (k=5)} & \textbf{Cache (k=10)} & \textbf{Cache (k=15)} & \textbf{Best Improvement}\\
\hline
MATH500 & 462 & 77.3\% & 80.3\% & 77.7\% & 80.3\% & 3.0\% \\
AIME2024 & 29 & 10.3\% & 13.8\% & 20.7\% & 13.8\% & 10.4\% \\
GPQA & 162 & 37.0\% & 38.9\% & 37.0\% & 34.6\% & 1.9\% \\
\hline
\end{tabular}
\caption{\label{tab:results}
Vanilla InftyThink Baseline vs.\ cache-augmented variants at different retrieval sizes.
}
\end{table*}

In our proposed model, we improve on vanilla InftyThink by storing useful strategies (lemmas) inside a cache (a vector database) after mean pooling. The mean pooling is done via the BGE-small embedder. At the end of each reasoning summarization step, the model returns the main strategy (lemma) used in that step along with the reasoning summary. Our method follows the same iterative summarization logic as InftyThink. However, unlike the vanilla version, our method retrieves the top-$k$ lemmas with the highest cosine similarity to the mean-pooled question embedding and displays them in the context window as optional hints. The cache is built at run time. Although our implementation is based on prompting, it can be extended to SFT or other fine-tuning approaches. Vanilla InftyThink risks losing details through repeated compression across numerous thinking steps; our approach partially mitigates this by using similarity-based cache retrieval. In this way, retrieved lemmas enter the model’s context and can steer subsequent reasoning toward more relevant strategies.

A specific transfer learning strategy was used for the AIME2024 dataset. Because the dataset contains only 30 questions, it is too small to build a mature, self-referential cache during runtime. Therefore, we applied a non-parametric transfer learning approach by utilizing the mature cache built during the MATH500 evaluation to answer the AIME2024 questions, capitalizing on the high domain overlap between the two datasets.


\section{Experiments and Results}
\subsection{Experimental Setup}

We selected Qwen-2.5-32B-Instruct as our primary inference engine. This model was chosen to align with the budgetary constraints and performance profiles favored in the original InftyThink literature. For the semantic retrieval component, we employed BAAI/bge-small-en-v1.5. This lightweight embedding model was responsible for vectorizing both the problem statements and the generated reasoning summaries, enabling the calculation of cosine similarity between a problem and the lemmas available within the cache.

The experiment was conducted across three datasets with distinct levels of difficulty and domain homogeneity: MATH500 (competition mathematics), AIME2024 (high-difficulty Olympiad mathematics), and GPQA Diamond (graduate-level multidisciplinary science). Pydantic was used to verify output formatting; questions where the model failed to produce outputs in the required format were excluded from the final analysis to ensure a fair comparison across control and treatment groups. Thus, the final testing sizes were $n=462$ for MATH500 (92.4\% retention), $n=29$ for AIME2024 (96.7\% retention), and $n=162$ for GPQA Diamond (81.8\% retention).

We established a Control Group (Vanilla InftyThink) and compared it against Treatment Groups utilizing cache sizes of $k=5$, $10$, and $15$. The reasoning generation was set to a temperature of 0.7 to align with the original InftyThink article, while the summarizer used a temperature of 0.0 to ensure concise lemma extraction. We utilized a top-$p$ value of 0.95. Due to cost constraints, each question was evaluated with a single stochastic run per condition; each run nonetheless consists of multiple iterative summarization steps, providing within-run structure. We therefore report one stochastic run model performance rather than averages over multiple independent runs.

\subsection{Analysis}

The quantitative results found above in Table~\ref{tab:results} indicate that the impact of semantic caching is highly dependent on the domain of the problem and the volume of retrieved context.

In the MATH500 dataset, the model demonstrated robust and stable improvements. The baseline accuracy of 77.3\% (Vanilla InftyThink) was surpassed by cache-augmented configurations. Specifically, both the $k=5$ and $k=15$ configurations achieved a peak accuracy of 80.3\%, representing a 3.0\% absolute improvement. The graph for MATH500 shows a U-shaped performance curve, suggesting that the model is resilient to varying context sizes in this domain, benefitting similarly from focused ($k=5$) and broader ($k=15$) retrieval.

The AIME2024 results showcased the highest volatility but also the most dramatic relative gains. Starting from a low baseline of 10.3\%, performance improved slightly with $k=5$ (13.8\%) and spiked with $k=10$ to 20.7\%, effectively doubling the baseline performance. However, performance dropped back to 13.8\% at $k=15$. While the small sample size ($n=29$) implies high variance, the peak at $k=10$ supports the transfer learning hypothesis, suggesting that reasoning patterns from MATH500 can transfer to harder AIME problems.

The GPQA Diamond dataset revealed limitations of the approach. The baseline accuracy of 37.0\% saw a modest improvement to 38.9\% with a focused cache ($k=5$). However, unlike the math datasets, increasing the retrieval size beyond $k=5$ caused degradation. At $k=15$, accuracy dropped to 34.6\%, performing worse than the no-cache baseline. This downward trend indicates that for heterogeneous domains, retrieving more information can introduce noise rather than signal.

\begin{figure*}[!b]
  \centering
  \includegraphics[width=\textwidth]{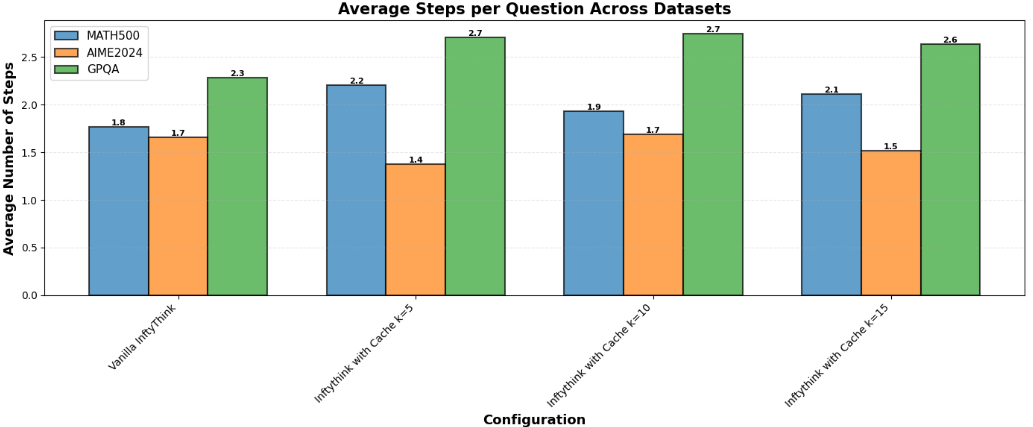}
  \caption{The Cost Difference Chart}
  \label{fig:cost-difference}
\end{figure*}

The Cost Difference Chart (Figure 2), which tracks the Average Steps per Question, provides insight into the efficiency of the cache. The cost difference is only statistically significant for the larger datasets (MATH500 and GPQA with $p$-value of 4\% and 2\% respectively), increasing the number of steps by an average of 0.26 steps (14\% increase) for MATH500 and 0.36 steps (16\% increase) for GPQA.


\section{Discussion}

Our results indicate that analytical reasoning improvement via embedding-similarity-based lemma retrieval exhibits mixed behavior. Across all cache window sizes, the retrieval-based model produces both \emph{fixes}, where questions answered incorrectly by the base model are answered correctly after lemma retrieval, and \emph{breaks}, where questions answered correctly by the base model are answered incorrectly after retrieval. The effectiveness of our approach, \emph{InftyThink with Cross-Chain Memory}, is strongly correlated with the problem domain. For the structured mathematical domains of MATH500 and AIME2024, cache-question similarity shows a positive correlation with performance. In contrast, for the domain-heterogeneous GPQA-Diamond dataset, higher similarity is primarily associated with breaks. We hypothesize that this divergence arises from the relative immaturity of lemma clusters in GPQA-Diamond, whose multi-domain structure of biology, physics, and chemistry limits the formation of coherent, task-specific lemma representations that are broadly applicable.

To understand the mechanics behind these outcomes, we performed a geometric analysis of reasoning trajectories in embedding space. A key observation is that lemma retrieval induces a significant directional shift in the model's reasoning relative to the base model’s average trajectory. This shift is consistently larger and more pronounced for summary chains, which we attribute to their semantically condensed structure and reduced representational variance. Interestingly, while lemma retrieval substantially alters reasoning trajectories, we observed no statistically significant directional differences across the different cache sizes of $k=5$, $10$, and $15$. This suggests that the presence of retrieved lemmas, rather than their quantity, is the primary driver of this directional change.

The core of our findings is the existence of \emph{directional attractors}. We constructed fix and break prototypes by mean-pooling the reasoning and summary chains corresponding to fix and break outcomes. We found that fix trajectories consistently align with the fix prototype, and break trajectories align with the break prototype in their average reasoning direction. The most striking discovery is that these two prototypes are statistically separable despite having an extremely high cosine similarity, approximately 98.8\% in both MATH500 and GPQA-Diamond. This indicates that correctness is not encoded in the overt semantic content of the reasoning itself but rather in the subtle \emph{directional change} of the reasoning trajectory relative to the no-cache baseline.

Multiple analyses support this geometric interpretation. Semantic manifold analysis reveals substantial overlap between fix and break chains, with purity levels below 58\%, confirming that cache effects act as small geometric shifts rather than clear semantic category shifts. Furthermore, our early divergence analyses show that the final outcome can be predicted with high accuracy from the initial tokens. For MATH500, early alignment of summary chains with the fix prototype results in a fix 75.6\% of the time, while early alignment with the break prototype yields a break with 82\% probability. Alignment energy analysis, which tracks cumulative distances to the prototypes, provides additional quantitative evidence that fix and break trajectories occupy geometrically distinct regions.

This behavior can be synthesized into a five-step reasoning pipeline: (1) reasoning is initialized from the input question; (2) relevant lemmas are retrieved via cache lookup; (3) a lemma-induced bias is injected into the early tokens of the response; (4) the reasoning path diverges toward either a fix or break directional attractor; and (5) the full reasoning trajectory aligns with the corresponding prototype.

Despite these insights, our study has several limitations. The analysis is based on a single LLM and a single embedding model, and the observed geometric phenomena may not generalize to all architectures. We evaluate each question with a single end-to-end run per condition. This run comprises multiple iterative summarization steps, which provides within-run structure, but it does not substitute for repeating the full pipeline under independent randomness. Consequently, our results characterize directional trends across questions, but they do not directly quantify per-question stability under repeated sampling. Apart from the initial refinement of the prompts explicitly specified in the Appendix \ref{app:prompt}, no additional prompt engineering, prompt tuning, or task-specific prompt modifications were applied. Finally, the inference order, which affects cache maturation, was not controlled. These limitations open several avenues for future work. The predictability of break attractors suggests that a "smarter" retrieval mechanism could be developed to filter out lemmas likely to cause breaks. Exploring alternative retrieval criteria beyond cosine similarity could also mitigate context pollution in heterogeneous domains. The finding that the final outcome can be predicted with high accuracy from the initial tokens could also be further explored for cheaper training methods of LLMs in the future.

\section{Conclusion}
In this work, we introduced \emph{InftyThink with Cross-Chain Memory}, a framework that integrates an embedding-based semantic cache into an iterative reasoning process. Our primary contribution is a detailed empirical and geometric characterization of how similarity-based retrieval influences LLM reasoning. We demonstrated that while this approach can improve accuracy on domain-specific benchmarks like MATH500, it can degrade performance on heterogeneous datasets like GPQA-Diamond by introducing distracting information.

Our main finding is that lemma retrieval actively steers the model's reasoning trajectory in embedding space, leading to consistent and predictable "fix" and "break" attractors. These attractors are geometrically separable and can be identified within the first few generated tokens, even though their high-level semantic content is nearly identical. This suggests that the success of memory-augmented reasoning hinges on controlling these directional biases rather than simply maximizing semantic relevance.

Similarity-based memory is a powerful but domain-sensitive tool with utility determined by the coherence of the knowledge domain. Additionally, the impact of retrieved context can be productively understood as a geometric perturbation in embedding space, a perspective that offers new ways to analyze and control LLM behavior. As next steps, we plan to leverage these findings to develop more sophisticated retrieval systems that can predict and actively avoid break attractors, paving the way for more robust and reliable self-improving reasoners.

\section*{Acknowledgements}

We thank Professor Siva Reddy for his valuable insights in shaping the direction of this work, and we are grateful to Megh Thakkar for helpful guidance and feedback throughout the project.

\bibliography{anthology,custom}
\bibliographystyle{acl_natbib}

\appendix
\section{Prompt Templates}
\label{app:prompt}

We report the exact prompt templates used for reasoning and summarization.

\subsection{Reasoning Prompt}

\begin{lstlisting}[basicstyle=\ttfamily\small,breaklines=true]
REASONING_INSTRUCTIONS = (
    "You are a careful competition-math solver. "
    "Show ALL arithmetic steps explicitly - don't skip calculations. "
    "For example, if you need to compute 5*7, write '5*7 = 35'. "
    "Use clear algebraic steps with intermediate results. "
    "Output plain text only. No JSON, no function calls, no summaries."
)
\end{lstlisting}

\subsection{Summarization Prompt}

\begin{lstlisting}[basicstyle=\ttfamily\small,breaklines=true]
SUMMARIZER_INSTRUCTIONS = (
    "You are a mathematical reasoning summarizer. "
    "Call 'summarize_and_decide' with:\n"
    "- summary_points: Key insights from reasoning (list of strings)\n"
    "- solution: ONLY if you have the FINAL EXACT ANSWER (just the number/expression)\n"
    "- main_trick: Key technique used \n"
    "If you see multiple values, list them in summary_points, NOT in solution\n"
    "ONLY set solution when reasoning shows COMPLETE calculation to final answer\n"
    "If reasoning says 'we need to compute X', DON'T set solution yet\n"
    "If unsure, leave solution as null and continue reasoning\n"
    "CRITICAL - LaTeX formatting rules for solution field:\n"
    "  * For infinity: ALWAYS write \\infty (NOT 'inf', 'infinity', or gibberish letters)\n"
    "  * For degrees: ALWAYS write complete X^\\circ (NOT incomplete X^ alone)\n"
    "  * For intervals: use proper notation with \\infty where needed\n"
    "  * ALWAYS complete LaTeX commands - NEVER leave dangling ^ or \\ characters\n"
    "  * Format examples: \\frac{a}{b} for fractions, \\sqrt{x} for roots, n^\\circ for degrees\n"
    "Common mistakes to avoid:\n"
    "- Setting solution to intermediate steps\n"
    "- Setting solution before verifying the calculation\n"
    "- Incomplete LaTeX: X^ without completing it (must be X^\\circ or X^{...})\n"
    "- Wrong infinity: random letters/numbers instead of \\infty\n"
)
\end{lstlisting}

\end{document}